\documentclass[12pt, a4paper]{article} 

\usepackage[utf8]{inputenc}
\usepackage[T1]{fontenc}
\usepackage{amsmath, amssymb}
\usepackage{graphicx} 
\usepackage{subcaption}    
\usepackage{hyperref}

\usepackage{booktabs}
\usepackage{geometry}

\usepackage[numbers,sort&compress]{natbib}
\usepackage{caption} 
\usepackage[ruled,vlined]{algorithm2e} %
\title{Phase transitions reveal hierarchical structure in deep neural networks}

\author{
Ibrahim Talha Ersoy\textsuperscript{1,}\footnotemark[1],
Andrés Fernando Cardozo Licha\textsuperscript{1,2,}\footnotemark[2],
Karoline Wiesner\textsuperscript{1,}\footnotemark[3]
}

\begin{document}

\maketitle

\noindent
\textsuperscript{1}University of Potsdam, Institute for Physics and Astronomy, Potsdam, Germany\\
\textsuperscript{2}Universidade Federal Fluminense, Instituto de Física, Niterói, Brazil

\footnotetext[1]{Corresponding author: talha.ersoy@uni-potsdam.de}
\footnotetext[2]{Email: andresfernando@id.uff.br}
\footnotetext[3]{Email: karoline.wiesner@uni-potsdam.de}

\vspace{0.5cm}

\begin{abstract}
    
\noindent Training Deep Neural Networks relies on the model converging on a high-dimensional, non-convex loss landscape toward a good minimum. 
Yet, much of the phenomenology of training remains ill understood. We focus on three seemingly disparate observations: 
the occurrence of phase transitions reminiscent of statistical physics, the ubiquity of saddle points, and phenomenon of mode connectivity relevant for model merging. We unify these within a single explanatory framework — the geometry of the loss and error landscapes.
We  analytically show that phase transitions in DNN learning are governed by saddle points in the loss landscape. 
Building on this insight, we introduce a simple, fast, and easy to implement algorithm that uses the L2 regularizer as a tool to probe the geometry of error landscapes. We apply it to confirm  mode connectivity in DNNs trained on the MNIST dataset by efficiently finding paths that connect global minima. We then show numerically that saddle points induce transitions between  models that encode distinct digit classes. 
Our work establishes the geometric origin of key training phenomena in DNNs and reveals a hierarchy of accuracy basins analogous to phases in statistical physics.

\end{abstract}

\section{Introduction}

Deep Neural Networks (DNNs) are the workhorse of modern machine learning. Training DNNs relies on the model converging on a high-dimensional and non-convex loss landscape toward a good minimum. Yet the reasons for their surprising efficiency in converging to good minima, as well as the question of how they encode information and compress data, remain poorly understood.
 A primary obstacle for developing an understanding of the behavior of DNNs is the complex, high-dimensional loss landscape, whose non-convex geometry critically guides optimization \cite{Choromaska2014, dauphin2014}. 

Recent work has shown that L2 regularization fundamentally alters landscape geometry by creating local minima \cite{mehta2022}, suggesting regularization's role extends beyond mere prevention of overfitting. 
Analytical frameworks from information-theory \cite{amari2016,watanabe2009}  and statistical physics \cite{Geiger2019,mehta2019,spigler2019jamming} provide insights into 
 phase transitions, learnability thresholds, and fundamental generalization properties of over-parameterized models. However, a unifying and practical principle for linking geometry to performance is still at large. We use the L2 regularizer to uncover such a principle. We focus on three phenomena associated with model convergence in DNNs: (1) the observation of 1st and 2nd order phase transitions at the onset of learning, akin to statistical physics \cite{ersoy2025,ziyin}; (2) the surprising ubiquity of saddle points on high-dimensional loss landscapes
\cite{Bray2007, dauphin2014} 
and their unclear role in optimization 
\cite{ge2015,jin2017}; 
and (3) the existence of almost flat paths between minima on the error surface (the so-called \emph{mode connectivity}) \cite{garipov2018loss,ainsworth2023gitrebasinmergingmodels} — paths that have recently gained much interest as a key to better generalization through model merging \cite{wortsman2022,ilharco2022}. 

L2 regularization is a standard method to prevent overfitting by penalizing large weights, thereby encouraging the model to prefer simpler solutions \cite{hoerl1970}. It is conceptually identical to an inductive bias \cite{xuhong2018} 
towards the model with all parameters equal to zero. L2 regularization is implemented by adding a quadratic penalty term to the loss function:
\begin{equation}  \mathcal{L}_\beta(\boldsymbol{\theta}) = E(\boldsymbol{\theta})+ \beta \|\boldsymbol{\theta}\|_2^2~,
 \label{eq:loss-standard}
\end{equation}
where $E(\boldsymbol{\theta})$ is the base loss function or error. $\boldsymbol{\theta}$ is the vector of all trainable parameters (weights and biases), and $\beta$ is the regularization strength. We refer to the space spanned by all parameters, with each point $\boldsymbol{\theta}$ as a model realization as the parameter space with the Euclidean norm of the parameter vector denoted by $r_0=\sqrt{\|\boldsymbol{\theta}\|_2^2}$, which defines the radial distance of a model realization $\boldsymbol{\theta}$ to the origin in parameter space or the distance to another models realization as $r_1=\sqrt{\|\boldsymbol{\theta} - \boldsymbol{\theta}_1\|_2^2}$. By error landscape we mean to the surface spanned by the corresponding error values over the parameter space and by loss landscape, the surface of the loss values. 
.  \\
Varying the regularization strength traces a path toward that trivial model (all parameters have value zero, up to noise) and is a (computationally much cheaper) analogue to the Information Bottleneck method \cite{tishby2000information,alemi2016deep}, which tracks the trade-off between data compression and relevant information preservation.
First and second-order phase transitions 
were initially identified at the onset of learning \cite{ziyin} where the two phases of random output and non-random output, respectively, are clearly separated by a jump in the error as a function of L2 regularization strength. 

 Our prior work established that applying L2 regularization to less trivially structured data reveals
 a hierarchy of phase transitions, each separating distinct regimes of model accuracy \cite{ersoy2025}. These transitions are accompanied by measurable geometric signatures in the error landscape's scalar curvature. This discovery of a multi-phase hierarchy, as opposed to a single transition, forces a re-evaluation of the error and loss landscape's geometry. We must now ask: Are these transitions local phenomena, or do they reflect a global organization? What is the precise geometry of these newly discovered boundaries and what do they reveal about models trained on complex data? 


Here, we move beyond phenomenology and the identity of geometric properties underlying the learning behavior of DNNs in general and the associated phase transitions in particular. We show analytically that phase transitions in DNN, indicated by discontinuous slopes in the error function, are universally governed by critical saddle points on the loss landscape that are associated with concave basin boundaries on the underlying un-regularized error landscape. 

Furthermore, we introduce a simple, easy to implement, and fast algorithm to explore the geometry of error landscapes, which we name \emph{Pathfinder}. We demonstrate the effectiveness and efficiency of this new \emph{Pathfinder} algorithm on DNNs trained on the MNIST dataset. For the MNIST dataset: (i) we identify hierarchical phase transitions between accuracy basins and find a striking correspondence between basins and specific digits; (ii) we confirm our hypothesis that 1st order phase transitions occur when the model passes a saddle point on the loss landscape, which must correspond to a concave transition point on the error landscape; (iii) we show that the concave transition points between accuracy basins form robust, extended sections, not mere points, corroborating a highly symmetric global organization of the error landscape; (iv) we track the paths connecting several optimal models, showing that there exist paths along which the error does not change substantially, and we demonstrate the computational efficiency and effectiveness of the \emph{Pathfinder} algorithm, which surpasses existing methods.
 
Our work establishes the geometric origin of key phenomenology in DNN training and demonstrates the existence of hierarchically ordered accuracy basins. These basins correspond to models in non-trivial error landscapes that are strictly analogous to phases in statistical physics.


\section{Results}

This work systematically dissects the geometric origins of phase transitions induced by L2 regularization in deep neural networks. We show that each transition is locally governed by the emergence of a saddle point, created when the L2 term amplifies a concave region of the underlying \textit{unregularized} error landscape. Building on this local mechanism, we demonstrate a global structure: the landscape is organized into concentric, hierarchically nested phases defined by weight norm. These phases are not isolated; they are intrinsically connected by low-error, flat pathways. To reveal these landscape features, we introduce the \emph{Pathfinder} algorithm (Methods, Algorithm~\ref{alg:pathfinder}), which probes and traverses transition boundaries.\\

\textbf{The geometric mechanism of regularization-induced phase transitions}
\label{sec:result-origin}
We begin by analytically establishing the local geometric mechanism underlying first-order phase transitions in the regularized loss landscape. As shown previously \cite{ziyin, ersoy2025}, such transitions involve a discontinuous shift in the global minimum as $\beta$ increases. We demonstrate that this shift occurs when the L2 term warps a concave region of the unregularized error landscape $E(\boldsymbol{\theta})$ into a saddle point. \begin{figure}[!ht]
    \centering
        \includegraphics[width=0.9\linewidth]{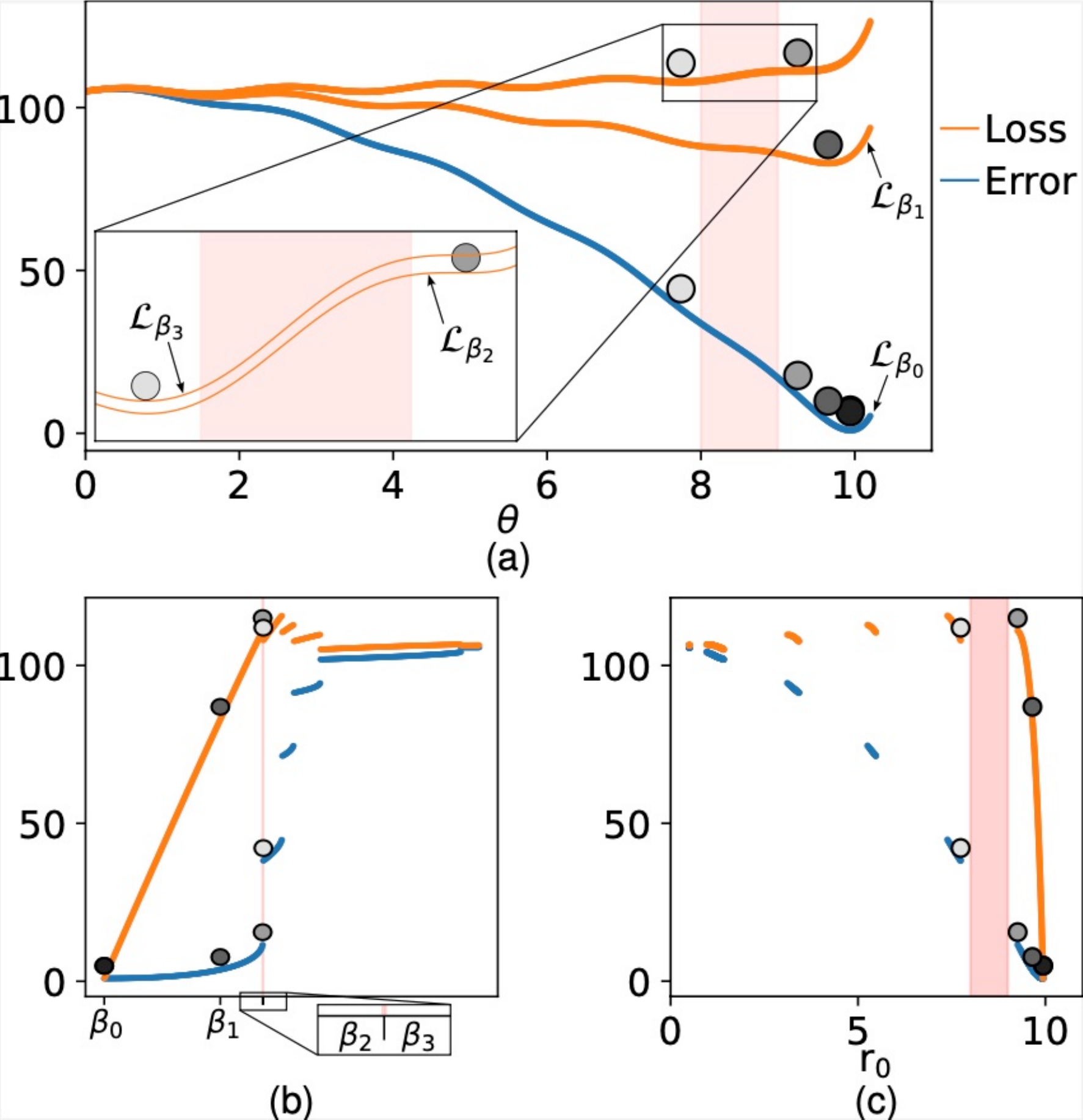}
    
    
     \caption{\textbf{Saddle-point mechanism: Transition is induced by L2 regularization.} Error $E(\boldsymbol{\theta})$ (blue) and loss $\mathcal{L}_{\beta}(\boldsymbol{\theta}) = E(\boldsymbol{\theta}) + \beta \|\boldsymbol{\theta}\|_2^2$ (orange) of a 1D toy model for four different regularization strengths $\beta$. Models (grey circles) positioned at respective loss minimum.
    (\textbf{a}) The error  remains the same for all $\beta$ while the loss  gets warped with increasing regularization strengths. Shifting $\beta$ from $\beta_0 \to \beta_2$,  the loss minimum and corresponding model  get slightly shifted towards the origin. Slightly shifting $\beta$, by $< 10^{-2}$, from $\beta_2 \to \beta_3$ causes a jump of loss minimum and model across the red-shaded concave section of the error and the corresponding maximum  of the loss, which becomes a saddle point in higher dimensions. 
    (\textbf{b}) Loss function $\mathcal{L}_\beta(\boldsymbol{\theta})$ and error $E(\boldsymbol{\theta})$ versus $\beta$ exhibit a continuous increase between $\beta_0$ and $\beta_2$, followed by a discontinuous transition between $\beta_2$ and $\beta_3$.
    (\textbf{c}) The same transition plotted against distance to the origin, $r_0$, which is equal to the root of the L2 norm, showing that the model jumps in parameter space between accuracy regimes.
    }
    \label{fig:mechanism}
\end{figure}
As illustrated in Fig.~\ref{fig:mechanism} for a 1D toy model, the regularized loss $\mathcal{L}_{\beta}(\boldsymbol{\theta})$ develops two separate minima along a concave segment of $E(\boldsymbol{\theta})$. At a critical $\beta_c$, these minima have equal loss, and beyond this point, the solution with lower L2 norm dominates, driving the transition. The resulting discontinuities in the loss   (Figs.~\ref{fig:mechanism}(b)) reproduce the known phenomenology of zero temperature phase transitions 
\cite{goldenfeld2018lectures} and confirm their geometric underpinning (Fig.~\ref{fig:mechanism}(c)). 
Mathematically, the formation of a saddle point and a corresponding phase transition follows directly from the structure of the  Hessian of the regularized loss:
\begin{equation}
\nabla^2 \mathcal{L}_{\beta}(\boldsymbol{\theta}) = \nabla^2 E(\boldsymbol{\theta}) + \beta I,
\label{eq:hessian}
\end{equation}
where $I$ is the identity matrix. The addition of $\beta$ shifts all eigenvalues upward, turning a direction with a negative eigenvalue on the error landscape into a saddle point on the loss landscape at the right $\beta$.
The critical regularization strength $\tilde{\beta}_c$ at which the local minima have the same loss value can be calculated \footnote{typically the transition will occur for larger $\beta_c \geq \tilde{\beta}_c$ values, depending on the learning rate, training algorithm and initialization}:
\begin{equation}
\tilde{\beta}_c = \frac{-\nabla E(\boldsymbol{\theta}) \cdot (\boldsymbol{\theta}-\boldsymbol{\theta}_{\text{ref}})}{2 \|\boldsymbol{\theta}-\boldsymbol{\theta}_{\text{ref}}\|_2^2},
\label{eq:transition_slope}
\end{equation}
where $\boldsymbol{\theta}_{\text{ref}}$ is the L2 reference point. In standard regularization, $\boldsymbol{\theta}_{\text{ref}}$ is typically the origin. However, in our \emph{Pathfinder} algorithm (See Methods), the reference point is chosen problem specific, probing and confirming the idea that these saddles emerge from intrinsic concave boundaries in $E(\boldsymbol{\theta})$.
In summary, L2 regularization transforms pre-existing concave boundaries in $E(\boldsymbol{\theta})$ into saddle points.  This provides a first-principles explanation for the hierarchical learning process in deep neural networks. In the following, we use the \emph{Pathfinder} algorithm to confirm this explanation for the MNIST data set. \\

\textbf{Probing saddle points and concave basin boundaries using the \emph{Pathfinder}}
\label{sec:result-pathfinder}
We empirically validate the above mechanism for phase transitions by applying the \emph{Pathfinder} algorithm to an MNIST classifier. Using annealing (see Methods
), we train multiple models along L2-controlled trajectories.

\begin{figure}[!ht]
\centering
\includegraphics[width=0.94\textwidth]{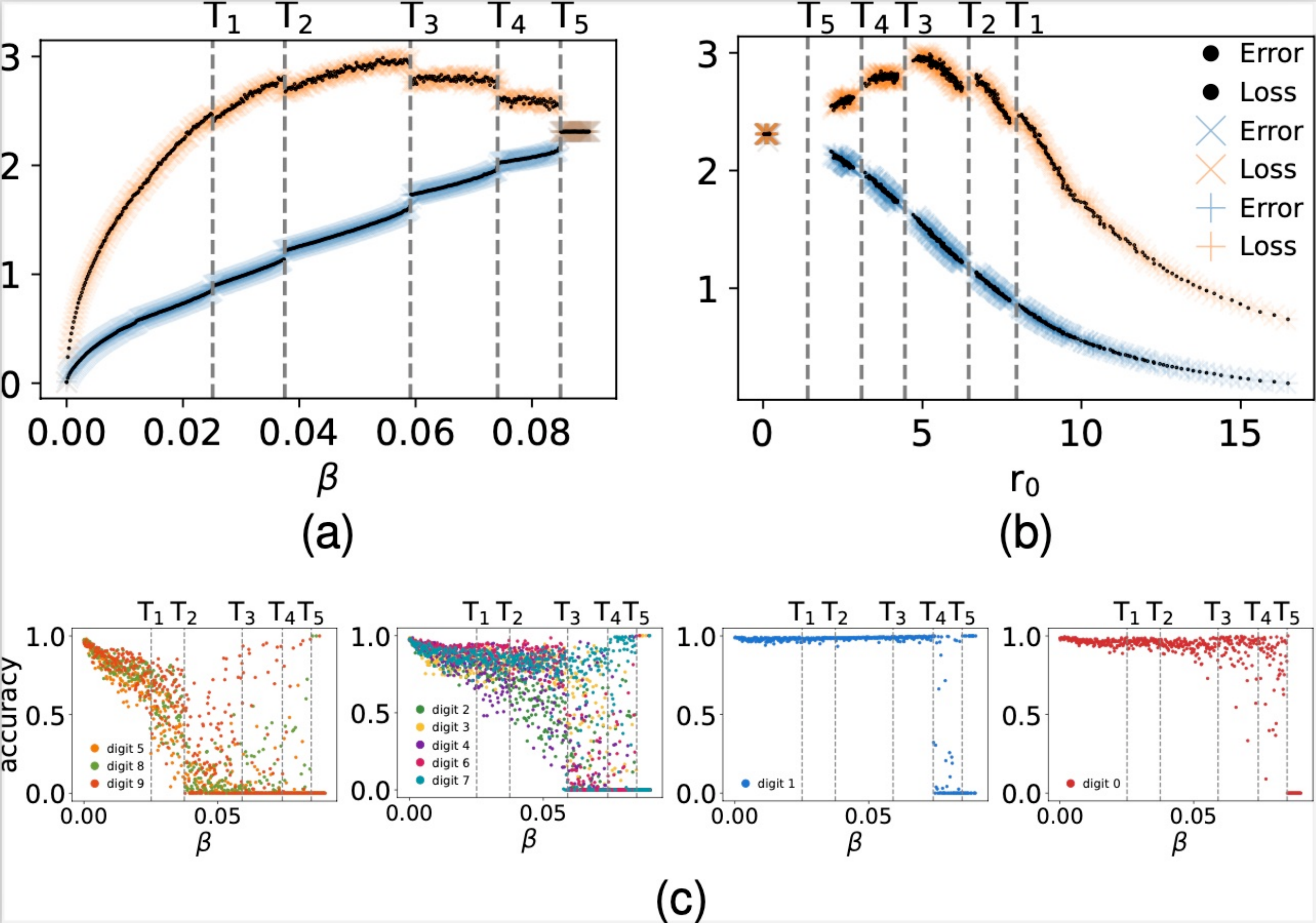}
%

%
\caption{
\textbf{L2 phase transitions correspond to (un)learning of features.} Error $E(\boldsymbol{\theta})$ and loss $\mathcal{L}_{\beta}$ of DNNs trained on the MNIST data, tracking three models $M_1 \left( \bullet \right)$, $M_2 \left(+\right) $ and $M_3 \left( \times \right) $ as a function of L2 regularization strength $\beta$, starting from distinct error / loss minima at $\beta = 0$. Phase transitions are marked $T_1$ to $T_5$.
The error $E(\boldsymbol{\theta})$ (blue) and loss $\mathcal{L}_{\beta}$ (orange) vs. $\beta$ curves (\textbf{a}) and  vs. distance to the origin in parameter space, $r_0$, (\textbf{b}) are identical for all model trajectories, indicating a high amount of symmetry in the error and loss landscapes.
(\textbf{c}) Per-digit accuracy reveals that specific features (and corresponding digits) are learned/forgotten at each transition $T_1$ to $T_5$.
}    \label{fig:global}  
\end{figure}

As shown in Fig.~\ref{fig:global}, we observe five phase transitions in error and loss (Fig.~\ref{fig:global}(a)), $T_1$ - $T_5$,  which each correspond to jumps in parameter space (Fig.~\ref{fig:global}(b)). There are further transitions in addition to these that become visible when we zoom in but here we only focus on the most pronounced ones. To study the global structure of the error landscape, we trained three converged models, $M_1$, $M_2$, $M_3$, at $\beta = 0$ and confirmed that they correspond to three different locations in parameter space. Using the \emph{Pathfinder} algorithm with the origin as reference point $\boldsymbol{\theta}_{\text{ref}}$ (see Eq.~\ref{eq:loss-pathfinder}), we  traced each model trajectory as a function of $\beta$. The resulting error and loss values are shown in Fig.~\ref{fig:global} but are indistinguishable as they are precisely on top of each other. This shows that phase transitions occur at fixed distances from the origin, regardless of model initialization. These numerical experiments establish that the landscape is organized into concentric, hierarchically nested phases defined by Euclidean norm of the parameter vector, i.e. the distance to the origin in the parameter space $r_0=\sqrt{\|\boldsymbol{\theta}\|_2^2}$. 
Furthermore, from Fig.~\ref{fig:global}(c) we can see that the phase transitions correspond to learning/forgetting events of specific digits. This finding links phases and their corresponding basins in the error landscape to the feature space at the class level, demonstrating  phase transitions in feature learning.

\begin{figure}[!ht]
\centering
\includegraphics[width=\textwidth]{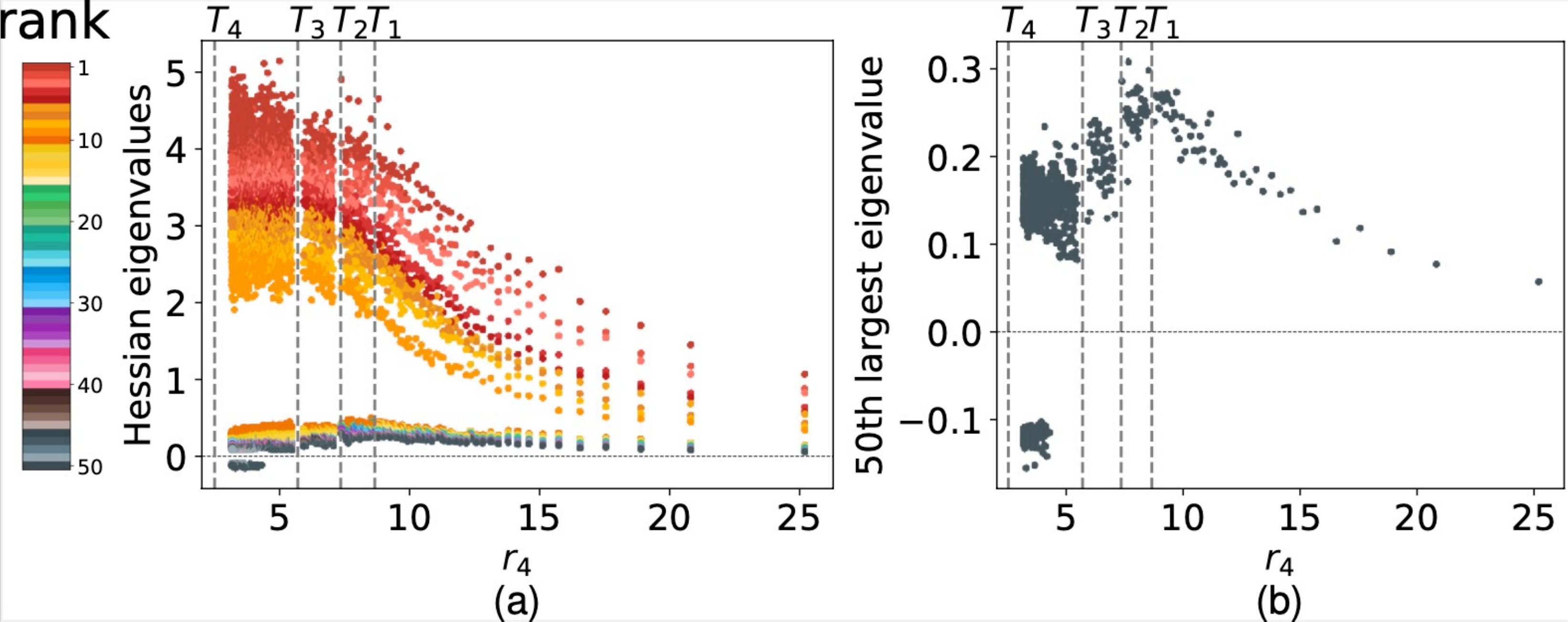}
\caption{
\textbf{Hessian spectrum of the error landscape confirms saddle-point formation.} Tracking converged model error $E(\boldsymbol{\theta})$ of DNN trained on the MNIST data as function of regularization strength, setting reference point $\boldsymbol{\theta}_{\text{ref}}$ (see Eq.~\ref{eq:loss-pathfinder}) to parameters of $T_4$. 
(\textbf{a}) Fifty largest  Hessian eigenvalues of the error  at converged model vs. distance $r_4$ to the reference point $T_4$. Most eigenvalues, except the nine largest ones remain close to zero. 
(\textbf{b})  50th largest eigenvalue plotted separately demonstrates a clear drop below zero as the transition point $T_4$ is approached}
\label{fig:pathfinder}
\end{figure}


To directly probe the concave boundaries of the error basins where the phase transitions take place, we set the reference point $\boldsymbol{\theta}_{\text{ref}}$ of the \emph{Pathfinder} algorithm to  parameter values near the saddle associated with transition $T_4$ by choosing a point between a trained model just before and one just after the transition, respectively. 
The results are shown in Figure~\ref{fig:pathfinder}. Negative curvature directions on the loss landscape emerge just before the phase transition takes place, while the bulk of the larger eigenvalues remains positive. This provides direct evidence of saddle-point formation and validates the geometric mechanism proposed in Sec.~\ref{sec:result-origin}. It also indicates that the model passes through a  canyon-like part of the error landscape.\footnote{We like to call this the William Tell effect, after the Swiss national hero's tale -- the unavoidable passage through a narrow lane.}\\

\begin{figure}[!ht]
\centering
\includegraphics[width=\textwidth]{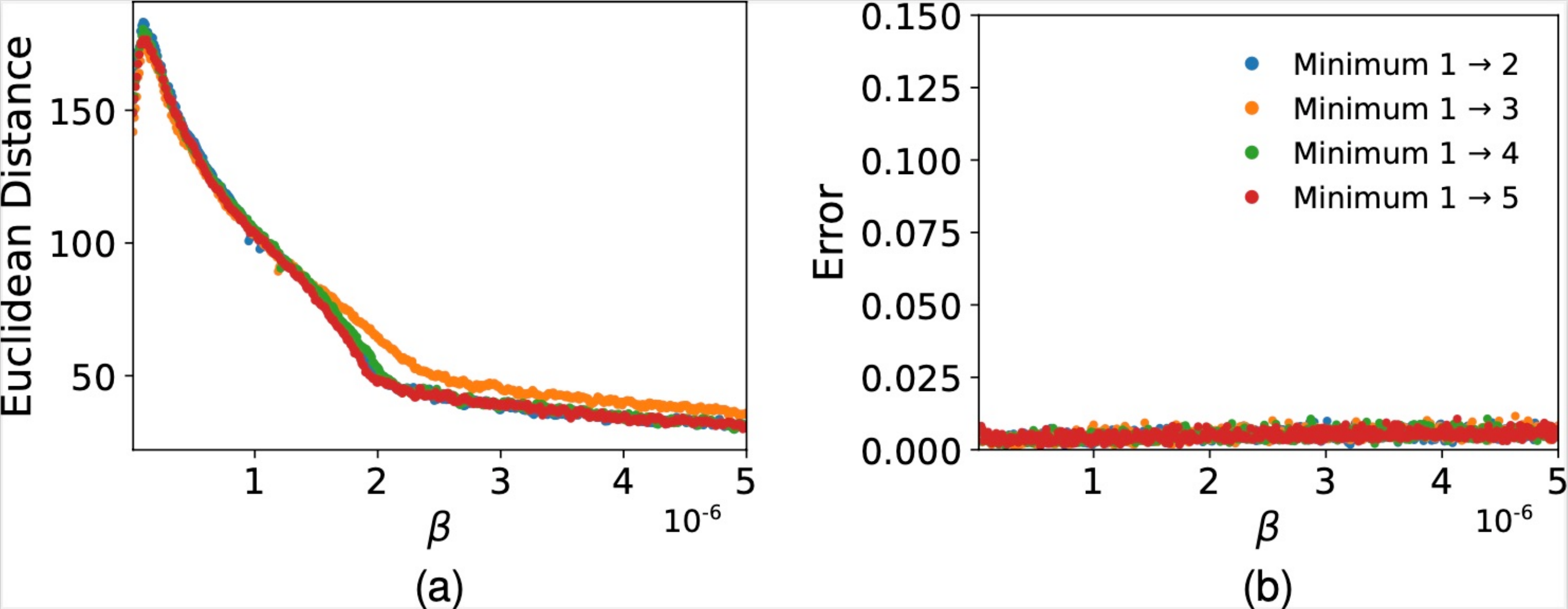}
\caption{
\textbf{Mode connectivity confirmed for MNIST data set.} Tracking converged model $M_1$ as function of regularization strength starting with $\beta = 0$, setting L2 reference point $\boldsymbol{\theta}_{\text{ref}}$ (Eq.~\ref{eq:loss-pathfinder}) to parameters of Minimum $M_2$, $M_3$, and $M_4$, respectively. 
(\textbf{a})  Euclidean distance to reference point  evolves nonlinearly with regularization strength $\beta$, confirming non-trivial path between minima. 
(\textbf{b}) Error of converged model remains nearly constant along the path between the Minima, with fluctuations two orders of magnitude smaller than the maximum error of $2.3$ (see Fig.~\ref{fig:global}).
}
\label{fig:connectivity}
\end{figure}

\textbf{Probing and confirming mode connectivity using the \emph{Pathfinder}
}
\label{sec:result-connectivity}
Finally, we examine whether independently discovered minimal-error solutions from the same accuracy phases are connected by continuous, low-error paths -- the so-called mode connectivity \cite{garipov2018loss}. To this end, we  randomly choose five optimal models (each trained with $\beta = 0$) that we confirm are not identical. We set the reference point $\boldsymbol{\theta}_{\text{ref}}$ of the \emph{Pathfinder} to the parameter values of model 1 (Minimum 1) and initialize the network with the parameters of the other four models, respectively, i.e. at the points in parameter space we call Minima 2 to 5. 
Fig.~\ref{fig:connectivity} shows the Euclidean distance between  initialized model and  reference point (Fig.~\ref{fig:connectivity}(a)) and the error (Fig.~\ref{fig:connectivity}(b)), each as a function of $\beta$. Note that $\beta$ only varies on the order of $10^{-6}$, which shows that the models are connected by flat, low-error paths (mode connectivity). This confirms that the landscape is not only phase-structured but also internally connected, with smooth, flat paths  between optimal models.

\section{Methods}

\textbf{Neural Network architecture and training}
Our experiments were conducted using the MNIST dataset of $28 \times 28$ pixel images of handwritten digits \cite{lecun2002gradient,deng2012mnist}. The  DNNs were trained with cross-entropy error \cite{murphy2012probabilistic,zhang2016understanding,goodfellow2016} to classify these handwritten images into digits ($0-9$). To ensure computational viability while mitigating finite-size effects associated with reduced parameter-space dimensionality, we used two fully connected hidden layers, each with $128$ neurons. The Sigmoid function was chosen as  activation function to ensure a smooth and differentiable error landscape \cite{cybenko1989,glorot2010}. ADAM optimization and a batch size of 64 were used to speed up convergence. To define convergence, training proceeded with a patience of 5 epochs. Various learning rates ranging from 0.0005 to 0.05 were tested. All simulations showed qualitatively similar results. For the  experiments, unless stated otherwise, the learning rate was $0.0015$, as it reduced deviations across different simulations of the same experiment, while still allowing the model to escape shallow local minima.\\

\textbf{Pathfinder algorithm for landscape exploration}
To systematically probing the geometry of the error surface, 
 we define an \emph{off-center} L2 regularizer, allowing us to direct the model toward any chosen reference point $\boldsymbol{\theta_{\text{ref}}}$ in parameter space:
\begin{equation}\label{eq:loss-pathfinder}
\mathcal{L}_{\beta,\boldsymbol{\theta}_{\textit{ref}}}(\boldsymbol{\theta}) = E(\boldsymbol{\theta}) + \beta \cdot \|\boldsymbol{\theta} - \boldsymbol{\theta_{\text{ref}}}\|_2^2.
\end{equation}
This formulation enables us to design trajectories connecting arbitrary  points in parameter space to the solution space  -- e.g. from an error minimum to the origin,  another minimum or any other point. We refer to this strategy as the \emph{Pathfinder} algorithm. 
This method also allows us to stear the network toward otherwise inaccessible regions of parameter space.  The algorithm proceeds by incrementally increasing the regularization strength $\beta$ and, for each $\beta$, training the model to minimize the  loss function of Eq.~\ref{eq:loss-pathfinder}. The resulting sequence of trained models traces a  path through the error landscape. It is worth noting that each step of the \emph{Pathfinder} is based on SGD, i.e. it finds paths in  regions where the actual learning dynamics takes place.
The algorithm can be used to trace model trajectories between critical points and, thus, to  identify phase transitions, to analyze flat versus sharp regions in the landscape, and  as an tool for exploring mode connectivity, all demonstrated in this paper. 
The full procedure is summarized in Algorithm~\ref{alg:pathfinder}.\\

\begin{algorithm}[H]
\caption{\emph{Pathfinder} Algorithm for Error Landscape Exploration}
\label{alg:pathfinder}
\KwIn{Initial parameters $\boldsymbol{\theta}_0$, error function $E(\boldsymbol{\theta})$, target parameters $\boldsymbol{\theta}_{\text{ref}}$, initial regularization strength $\beta_0 = 0$, maximum $\beta_{\text{max}}$, increment $\Delta\beta$, convergence threshold $\epsilon$}
\KwOut{Final parameters $\boldsymbol{\theta}^*$ close to $\boldsymbol{\theta}_{\text{ref}}$}
$\boldsymbol{\theta}_{\text{current}} \leftarrow \boldsymbol{\theta}_0$ \tcp*{Start from initial parameters}
$\beta \leftarrow \beta_0=0$ \tcp*{Initial regularization strength}

\While{$\|\boldsymbol{\theta}_{\text{current}} - \boldsymbol{\theta}_{\text{ref}}\|_2^2 \geq \epsilon$ and $\beta \leq \beta_{\text{max}}$}{
    \tcp{Train to convergence with loss $L(\boldsymbol{\theta}) = E(\boldsymbol{\theta}) + \beta\|\boldsymbol{\theta} - \boldsymbol{\theta}_{\text{ref}}\|_2^2$}
    $\boldsymbol{\theta}_{\text{current}} \leftarrow \text{TrainToConvergence}(\boldsymbol{\theta}_{\text{current}}, E(\boldsymbol{\theta}) + \beta \|\boldsymbol{\theta} - \boldsymbol{\theta}_{\text{ref}}\|_2^2)$
    
    $\beta \leftarrow \beta + \Delta\beta$ \tcp*{Increase regularization and continue}
}

\Return $\boldsymbol{\theta}_{\text{current}}$ \tcp*{Return final parameters}
\end{algorithm}




%

\begin{figure}[!ht]
\centering
\includegraphics[width=\textwidth]{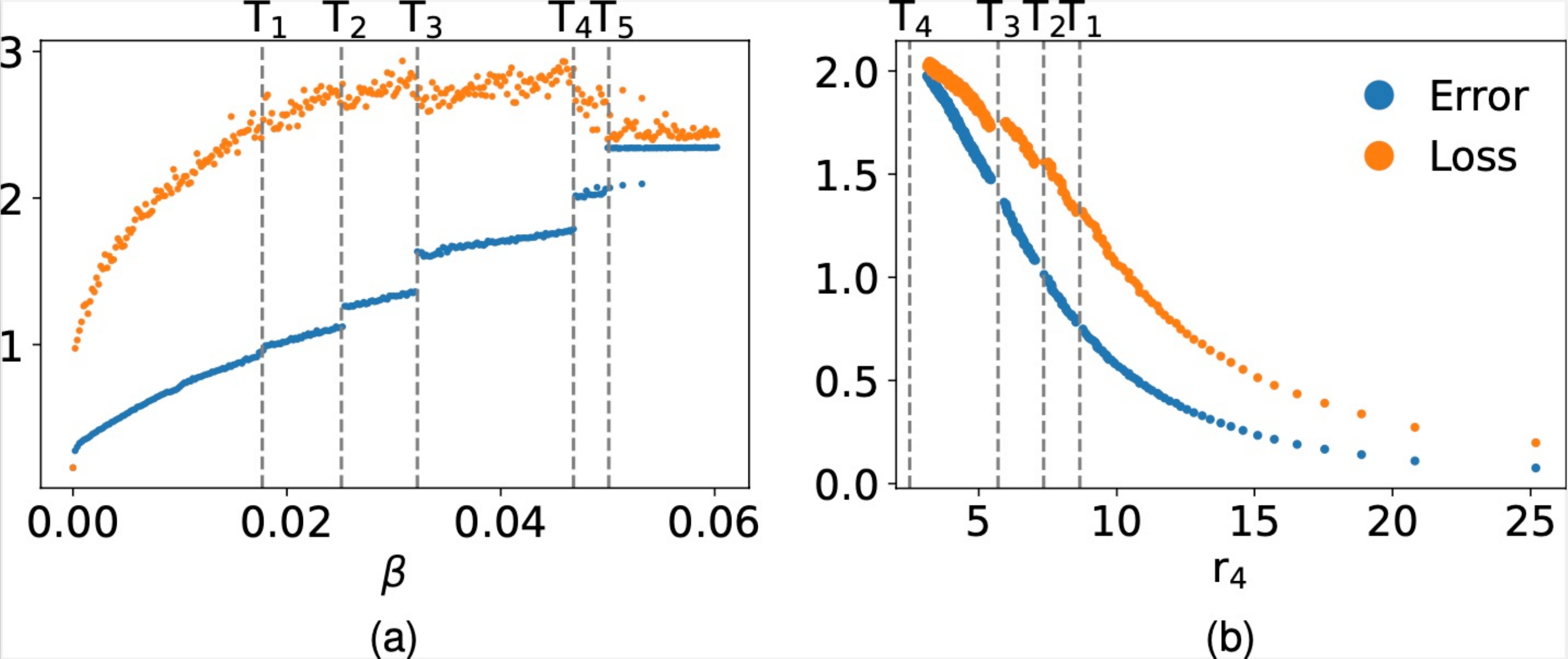}
\caption{
\textbf{Identified Transitions are Robust w.r.t. Learning Rate and L2 reference point.} Same experiment as in Fig.~\ref{fig:global} but with faster learning rate ($0.015$)  
(\textbf{a}) and different L2 reference point $\boldsymbol{\theta}_{\textit{ref}}$ (\textbf{b}). Transitions in model error (blue) and loss (orange) as a function of regularization strength $\beta$ with  learning rate $0.015$ are identical to those obtaiend with learning rate $0.0015$ (Fig.~\ref{fig:global}), indicating they are not optimization artifacts.
(\textbf{b}) Shifting the L2 regularization reference $\boldsymbol{\theta}_{\textit{ref}}$ to the fourth transition ($T_4$) maintains the core phase structure, indicating the robustness of the transitions with respect to experimental setup (learning rate $0.0015$).
}
\label{fig:robustness}
\end{figure}

\textbf{Robustness of Trajectories.} To ensure that the observed transition phenomenology correspond robust geometric features of the error landscape and not artifacts of a specific optimization setting, we performed robustness tests. We verified that trajectories generated with different learning rates (Fig.~\ref{fig:robustness}a) and with different endpoint configurations (Fig.~\ref{fig:robustness}b) consistently crossed transition boundaries at the same critical Euclidean norm. The reproducibility of these transitions across varied optimization conditions confirms their origin in the underlying landscape geometry.\\

\textbf{Reproducibility and Data Availability}

All experiments were conducted using publicly available data and open-source tools. The MNIST dataset is publicly accessible at \url{http://yann.lecun.com/exdb/mnist/}. 

The full source code used for model training, \emph{Pathfinder} trajectory generation, and Hessian analysis is available at:

\url{https://github.com/andresfernando-git/Neural-Network-paper.git}

Code, processed data, trained models, and additional materials will be made available in a public data repository after peer review is complete.

\section{Discussion}


Our work connects three seemingly disconnected and not fully explained phenomena in deep learning to the geometry of the loss and error surfaces: (i) the observation of phase transitions in regularised learning, (ii) the ubiquity of saddle points, and (iii) mode connectivity. 
We demonstrate that the error landscape of deep neural networks is organized into a hierarchical structure of ``accuracy basins''. By re-purposing the L2 regularizer, $||\boldsymbol{\theta} - \boldsymbol{\theta}_{ref}||_2^2$, as a systematic probe, we showed that these basins are hierarchically stacked and separated by concave  boundaries. During regularisation, models pass these boundaries and cause a phase transition in the error.   We provided a mechanistic, geometric explanation for these phenomena by proving that the regularizer induces saddle points in the loss landscape at the concave boundaries on the underlying error landscape. With these results, we move beyond the empirical observation of phase transitions \cite{ziyin, ersoy2025} to a formal geometric theory, which connects the physics of phase transitions to the dynamics of deep learning.
Furthermore, we demonstrated that the regulariser as a probe is a powerful general tool -- which we call \emph{Pathfinder}-- for exploring mode connectivity and other geometric properties of the error surface.\\

\textbf{A Computationally Efficient Pathfinder for Mode Connectivity and Beyond}
Our shifted-L2 method (\emph{Pathfinder}) introduced here provides a distinct approach to exploring loss landscapes including the connectivity of solutions, a topic of significant recent interest \cite{garipov2018loss,ainsworth2023gitrebasinmergingmodels}. The predominantly used methods for finding low-error paths between modes are either optimization-based, training parameterized curves between models \cite{garipov2018loss}, or symmetry-based,  finding permutation-equivalent models connected by linear paths \cite{ainsworth2023gitrebasinmergingmodels}.
Our approach offers key practical benefits. It is conceptually simpler and  computationally more efficient than optimization-based algorithms. Techniques like the Bezier curve method \cite{garipov2018loss} require a costly secondary optimization loop, training a ``control model'' to minimize the average loss along the path --a process involving repeated model blending gradient calculations through this blend. In contrast, the \emph{Pathfinder} method traces a path by performing multiple standard training runs with a simple, modified L2 loss (see Table~\ref{alg:pathfinder}). While both methods require multiple runs, our algorithm uses a conventional, highly optimized training procedure entirely avoiding the complexity and overhead of a custom path-optimization meta-loop.
Furthermore, the \emph{Pathfinder} method is more general than purely symmetry-based methods. It can discover connecting paths governed by the landscape's fundamental hierarchical organization, not just those arising from permutation symmetry. 

As we demonstrated on MNIST, this allows us to  confirm mode connectivity and, more interestingly, to directly reveal the hierarchical structure of the error landscape. Specifically, we observed five distinct phase transitions where features are progressively gained / lost. Crucially, these transitions correspond to meaningful model changes: some transitions cause the gain / loss of multiple digit classifications simultaneously, while others affect only a single, specific digit. This demonstrates that our method can identify transitions between functionally distinct models, directly linking features to the landscape's intrinsic phase structure.\\


\textbf{Statistical physics of deep learning}
Our findings position neural networks firmly within the realm of physical systems amenable to analysis by statistical physics concepts and methods. The observed phase transitions are not merely metaphorical; they are quantitative phenomena with identifiable geometric signatures. This connection allows us to move beyond analogy and apply the rigorous machinery of statistical physics directly to deep learning.
For instance, the global organization we uncovered suggests that methods like topological data analysis could be used to classify basin structures and explain phenomenology in the training using the well established knowledge of critical phenomena. Crucially, our analytical proof (Eq.s~\ref{eq:hessian},\ref{eq:transition_slope}), that the regularizer creates saddle points at concave sections of the error landscape corresponding to basin boundaries, explains the existence of phase transitions in deep learning. We thus have moved from statistical-physics phenomenology to a precise geometric explanation.
This perspective treats the learning process not as a mere optimization task but as a physical process exploring a structured landscape, where the shifted L2 strength $\beta$ acts as a thermodynamic parameter tuning the system between phases.
This mapping provides a mechanistic explanation for several empirical observations in deep learning. For example, the efficacy of transfer learning \cite{kirkpatrick2017overcoming} can be understood as initializing a model in a broad, high-performance basin from which it can easily fine-tune to a related task. Similarly, the benefits of curriculum learning \cite{wang2021survey,bengio2009curriculum} may stem from guiding the model through a path that avoids sharp, hard-to-traverse geometric barriers.\\


\textbf{A practical analogue to the Information Bottleneck Method}
The loss function (Eqs.~\ref{eq:loss-standard} and \ref{eq:loss-pathfinder}) represents a trade-off between model complexity (L2 term) and model performance (error term) very similar to the Information Bottleneck (IB) method \cite{tishby2000information} and its variational approximation, the ``deep variational IB'' \cite{alemi2016deep}. Both IB and variational IB are based on mutual information which remains very difficult and expensive to estimate \cite{pmlr-v80-belghazi18a,goldfeld2020information, michael2018on,noshad2019scalable}. Therefore, practical implementation in the context of deep learning remains impossible for IB and prohibitively expensive for variational IB. Furthermore, variational IB requires a change of the existing architecture as it introduces a latent so-called bottleneck layer. Our work introduces a computationally cheap and practical analogue to the IB method that can, in theory, be applied on any parametric model. We hypothesize that a similar basin hierarchy will emerge in various architectures beyond feed-forward networks, in most non-convex optimization methods. This makes our method a potent diagnostic tool, allowing us to identify geometric features by detecting change-points in the L2 paths. On MNIST, the five observed transitions create a taxonomy of feature importance: the sequence in which the digits are lost (with some grouped and others isolated) reveals the model's implicit hierarchy of features. By identifying which features become accessible or degraded at which regularization strength, we can begin to reverse-engineer the network's internal representation, providing a first principled step toward explaining what the networks truly learn.\\


\textbf{Limitations of the Pathfinder method}
Despite the novelty and effectiveness of our perspective and methodology, it is subject to certain conditions. The method relies on the assumption that the error landscape possesses a degree of smoothness and connectivity that allows for meaningful paths to be traced. Furthermore, the computational advantage, while significant compared to IB calculations, still requires multiple optimization runs at different regularization strengths. The optimal range for $\beta$ is task- and architecture-dependent, presenting a practical hyperparameter search challenge.
Our numerical results are currently confined to feed forward architectures, where we know that the L2 scheme produces first order phase transitions in deep- and second order transitions in one hidden layer- neural networks. A critical open question is how this hierarchy manifests in other architectures (e.g., CNNs, Transformers).  We hypothesize that the underlying principle—concave basin boundaries giving rise to regularizer-induced saddle points and associated phase transitions—should hold for any sufficiently complex, non-convex model space. Finally,  the precise relationship between the L2 path and the path taken by a standard stochastic optimizer during training warrants further investigation.\\


\textbf{Conclusions and future directions}
This work opens several compelling avenues for future research. Theoretically, it invites a concerted effort to apply statistical physics methods to quantitatively characterize error landscapes, their role in the learning dynamics, and the role of criticality in information processing. Practically, the shifted L2 probe (\emph{Pathfinder}) provides a new diagnostic framework for analyzing model behavior and for designing better training schemes. From an interpretability standpoint, it lays the groundwork for reverse-engineering the hierarchical features a network acquires. Not least, a better understanding of the error-landscape geometry will be key for developing better strategies and methods for continuous learning, transfer learning and model merging. In conclusion, by re-conceptualizing a standard regularizer as a probe, we have opened the black box of DNNs a little further. The physics analogy allowed us to understand them as systems with a a comprehensible internal organization.
This not only deepens our theoretical understanding of Deep Learning but also provides a practical path toward more robust, interpretable, and efficient machine learning.

\section*{Acknowledgements}
We are grateful to Björn Ladewig for the insightful discussions. A.F.L.C. acknowledges the DAAD for scholarship Research Grants - One-Year Grants for Doctoral Candidates (7693452) to the University of Potsdam. 

\section*{Author Contributions}
I.T.E. and K.W. designed the study, I.T.E. conceptualized the idea of the \emph{Pathfinder}. I.T.E. and A.F.L.C. designed the methodology. A.F.L.C. conducted the numerical experiments and produced the figures.  All authors discussed the results and contributed to the writing of the manuscript.

\section*{Competing Interests}
The authors declare no competing interests. 
\bibliography{references}

\end{document}